\documentclass[conference]{IEEEtran}
\IEEEoverridecommandlockouts
\usepackage{cite}
\usepackage{amsmath,amssymb,amsfonts}
\usepackage{algorithmic}
\usepackage[ruled,vlined,linesnumbered]{algorithm2e}
\usepackage{mathptmx}
\usepackage{graphicx}
\usepackage{textcomp}
\usepackage{xcolor}
\usepackage{comment}
\def\BibTeX{{\rm B\kern-.05em{\sc i\kern-.025em b}\kern-.08em
    T\kern-.1667em\lower.7ex\hbox{E}\kern-.125emX}}
\begin{document}

\title{Towards Enabling Dynamic Convolution Neural Network Inference for Edge Intelligence\\}

\author{
    \IEEEauthorblockN{
    Adewale Adeyemo\IEEEauthorrefmark{0},
    Travis Sandefur\IEEEauthorrefmark{0},
    Tolulope A. Odetola\IEEEauthorrefmark{0},
    Syed Rafay Hasan\IEEEauthorrefmark{0},
    }
    \IEEEauthorblockA{
    \IEEEauthorrefmark{0}Department of Electrical and Computer Engineering, Tennessee Technological University, Cookeville, TN 38505\\
  }
}

\maketitle

\begin{abstract}
Deep learning applications have achieved great success in numerous real-world applications. Deep learning models, especially Convolution Neural Networks (CNN) are often prototyped using FPGA because it offers high power efficiency, and reconfigurability. The deployment of CNNs on FPGAs follows a design cycle that requires saving of model parameters in the on-chip memory during High level synthesis (HLS). Recent advances in edge intelligence requires CNN inference on edge network to increase throughput and reduce latency. To provide flexibility, dynamic parameter allocation to different mobile devices is required to implement either a predefined or defined on-the-fly CNN architecture. In this study, we present novel methodologies for dynamically streaming the model parameters at run-time to implement a traditional CNN architecture. We  further  propose  a library-based approach to design scalable and dynamic distributed CNN inference on the fly leveraging partial-reconfiguration techniques, which is particularly suitable for resource constrained edge devices. The proposed techniques are implemented on the Xilinx PYNQ-Z2 board to prove the concept by utilizing the LeNet-5 CNN model. The results show that the proposed methodologies are effective, with classification accuracy rates of 92\%, 86\% and 94\% respectively\footnotetext{This paper has been submitted for publication in ISCAS, 2022}. 

\end{abstract}

 \vspace{-2.5mm}


\section{Introduction}
Deep learning models have found acceptance in various field due to its improved performance \cite{zhang2021deepslicing}, \cite{adeyemo2021security}. To achieve fast prototyping, high energy efficiency and low overhead, FPGAs are used to accelerate deep learning models especially, Convolutional Neural Networks (CNNs) \cite{odetola2021sowaf, shawahna2018fpga}. CNNs are a popular deep learning architecture for image classification due to their ability to extract many complex high-level features required for object classification \cite{stahl2021deeperthings}. As data grows due to recent improvements in smart devices, there is a demand for more processing power, which is moving the core of computation from the cloud to the network's edge \cite{li2018learning}. 

Edge intelligence \cite{zhou2019edge} pushes deep learning tasks from the cloud to the edge, enabling a variety of distributed, low-latency, and dependable intelligent applications\cite{wang2020convergence}. However, as the number of queries to be handled increases, a single edge node may be unable to handle such resource-intensive deep learning applications. An alternate solution is to use multiple cooperative edge devices to perform the CNN inference task in a distributed and cooperative manner, a phenomenon known as horizontal collaboration \cite{adeyemo2021security}. Horizontal collaboration entails partitioning and assigning trained CNN models to multiple edge nodes to accelerate deep learning computations \cite{odetola2021feshi}. The allocation of these CNN layers across multiple end nodes would require dynamic allocation of parameters to the model layers. 

 DeeperThings, proposed in \cite{stahl2021deeperthings},  partitions fully-connected layers, as well as feature and weight intensive convolutional layers, to serve a wide range of CNN inference tasks. By integrating feature and weight partitioning with a communication-aware layer fusion approach, holistic optimization across layers was achieved, allowing memory and computation demands to be optimized simultaneously. In \cite{farley2021memory}, prior fusing strategies are extended by the authors to enable for two groups of convolutional layers to be fused and tiled separately. This method saves overhead by reusing data and shrinks the memory footprint even more. MoDNN, a local distributed mobile computing system that executes deep learning computations on mobile platforms while balancing worker node workloads and reducing data delivery time was introduced in \cite{mao2017modnn}. Each neural layer is divided into slices, which are then executed layer by layer to increase parallelism and reduce memory footprint. Experiments showed that MoDNN has a huge potential for mobile platforms in DNN applications. In \cite{alwani2016fused}, dataflow across convolutional layers was investigated. Authors are able to fuse the processing of multiple CNN layers by rearranging the order in which input data is received on the chip, allowing intermediate data to be cached between CNN layer evaluations. The current state of art methodologies for horizontal collaboration assume that the CNN architecture and trained parameters are  known to each participant beforehand \cite{wang2020convergence}. Since, providing parameters in advance may lead to security issues, hence it is important to dynamically provide parameters. Secondly, dynamic allocation of resources can help the usage of same CNN for multiple training dataset. Another concern in modern distributed CNN is the possibility of dynamic allocation of resources to implement a CNN layer in run-time, especially in a resource constrained device. Hence there is a growing need to address these two problems and this paper is an attempt towards their solution.

In this paper, a novel method of CNN acceleration on FPGAs that dynamically stream in model parameters at runtime rather than have them in on-chip memory beforehand was proposed. We also provide a proof-of-concept method for dynamic resource allocation leveraging the partial-reconfigurability of FPGAs via the Vivado PR library. This methodology dynamically design a CNN layer which can also alleviate the scalability issues when large CNNs are deployed on resource constrained FPGA based edge devices. The design is implemented on the Xilinx PYNQ-Z2 board with a LeNet-5 model trained on MNIST dataset.

The remainder of the paper is structured as follows: Section
II describes the Background, Section III describes Methodology proposed to dynamically stream in weights to CNN layers on-the-fly, section IV describes our experimental results  and Section V concludes this paper.

\section{Background}


\subsection{Design Flow for the Deployment of CNN models to Xilinx FPGA}
In deploying an hardware to Xilinx FPGA, the design flow are in two stages which are the software and hardware. The software operation involves preparation of dataset, training of model and weight extraction. The hardware design involves the $C^{++}$ layer by layer abstraction which is done using the Vivado HLS, block design which is done using Vivado and deployment of the CNN on the FPGA which is validated using python overlay on the Xilinx PYNQ-Z2. 


\subsection{Partial Reconfiguration (PR)}

Partial Reconfiguration (PR) is a design technique for FPGAs that enables dynamic module changes in an active design using the Xilinx Vivado tool. The design cycle mandates the production of a large number of configurations, which results in entire bitstreams for each configuration and partial bitstreams for each reconfigurable module. \cite{kao2005benefits}. Some of the benefits of PR include the ability to reconfigure during runtime, as well as reduce size, weight, power, and cost \cite{irmak2021increasing}.

 \begin{figure}[]
\centerline{\includegraphics[width=.5\textwidth]{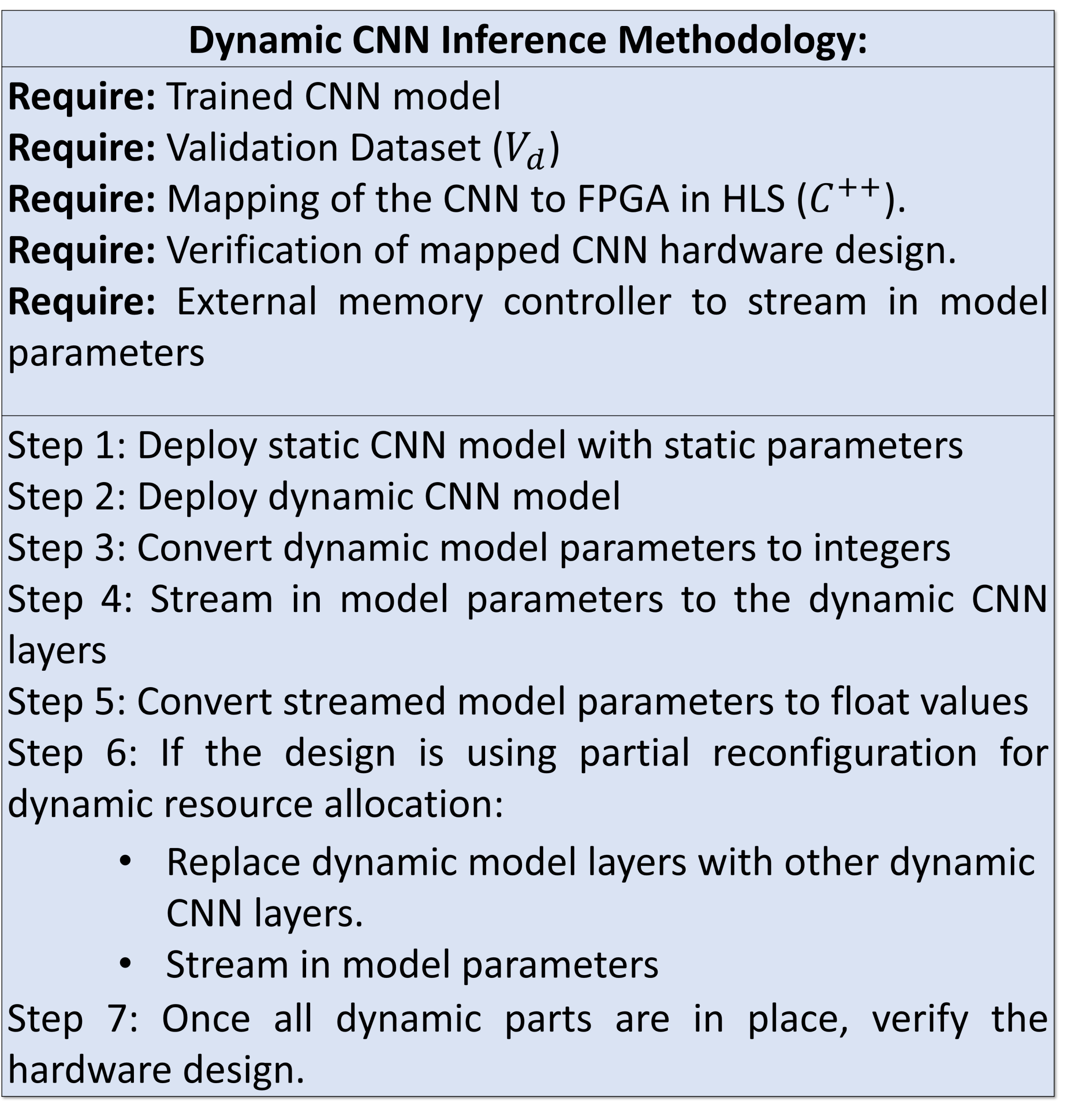}}
\caption{General Dynamic CNN inference Step-by-Step Methodology for dynamic parameter allocation and dynamic resource allocation}
\label{image20}
\end{figure}
\section{Dynamic CNN Inference}

 \begin{figure}[]
\centerline{\includegraphics[width=.5\textwidth]{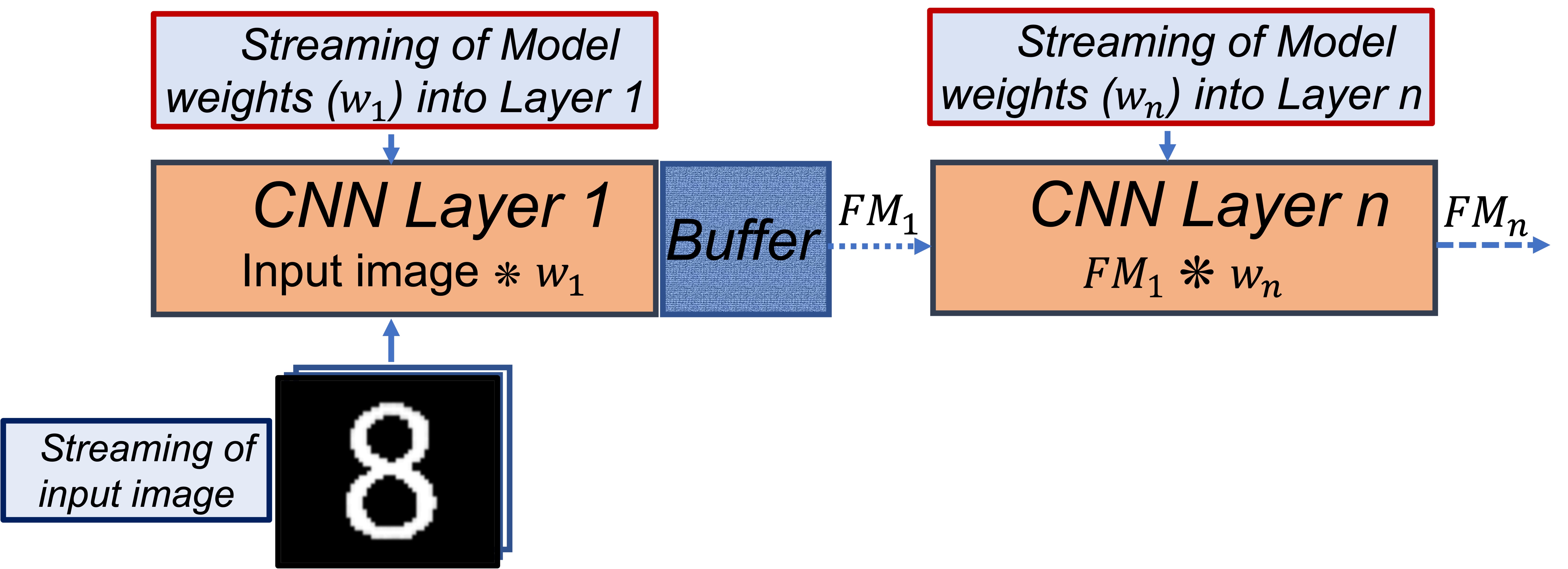}}
\caption{Overall illustration of how the proposed design achieves dynamic parameter allocation. In the first CNN layer, the parameters and input image can be streamed. All other layers can be dynamically provided with parameters, as shown, and feature maps (FM) are normally passed to the subsequent layer(s). For DASP parameters for all the layers are streamed in, for DAPP parameters are streamed in for selected layers to avoid bandwidth constraints (rest of the layers remain static on the FPGA).}
\label{image11}
\end{figure}



In this work, we propose dynamically streaming in parameters of CNN layers at runtime (see Fig. \ref{image20}). The general methodology is illustrated in Fig. \ref{image20}. Overall there are two major approaches proposed, (1) dynamic parameter allocation and (2) dynamic resource allocation. Steps 1 to 5 in Fig. \ref{image20} shows the general requirement for both of them and Step 6 is specific to the dynamic resource allocation. It is to be noted that we provided two variants of Dynamic Parameter Allocation technique, namely (1) Streaming in parameters for all the layers and (2) Bandwidth efficient steaming of the parameters of only few layers (and the rest of the CNN layers remain static on FPGA), further details are provided in Sections \ref{SectionA} and \ref{SectionB}. Dynamic Resource Allocation is made possible using Library-Based PR along with dynamically streaming in the parameters (details are provided Section \ref{SectionC}). 

\subsection{Dynamic Allocation of Streamed Parameters (DASP): Streaming of parameters into all the CNN layers in hardware design Phase} \label{SectionA}
The steps below describe the DASP methodology following the general methodology shown in Fig. \ref{image20}
\begin{itemize}
\item No static model parameters are required for this technique.
\item Vivado HLS design ensures that model parameters are dynamically streamed for all layers.
\item The hardware design is synthesized using Vivado before being deployed to the FPGA (Xilinx- PYNQ Z2 in our case). In our methodology, we utilize the processing subsystem (PS) to control the dynamic streaming of the model parameters. Necessary data type parsing needs to be performed to accommodate the requirements of FPGA design. All the streamed in data are required to be in integer format for dynamic streaming in of parameters in Xilinx's design tools (step 3). Since model weights obtained after training are often in floats, the float values need to be converted to integers. After being streamed in, the parameters are converted back to float values to reduce the quantization accuracy loss (step 5). The bit-stream file obtained after hardware design on Vivado is deployed to the Xilinx PYNQ-Z2 board for validation, and the model parameters are dynamically streamed in using the PS of Zynq FPGA to perform the image classification (see step 7). This method provides a proof of concept strategy to dynamically allocate CNN parameters to enable one type of dynamic CNN inference.  
\end{itemize}

\subsection{Dynamic Allocation of Partial Parameters (DAPP): Dynamic streaming on parameters to select layers of the CNN in hardware design Phase} \label{SectionB} 
As a follow up to the methodology (DASP) described above, we note that streaming of network parameters requires enormous data bandwidth. In order to efficiently utilize the data bandwidth, we  provide a proof of concept strategy to dynamically allocate CNN parameters to only selected layers of the CNN while other layers of the CNN have their parameters saved in the on-chip memory. The overall methodology for DAPP is described in the steps below as shown in Fig. \ref{image20}. 
\begin{itemize}
\item Static model parameters are obtained and saved after training of the model. 
\item Only selected (partial) model parameters are dynamically streamed utilizing Vivado HLS tool. This helps in selecting target layers with large weights to utilize both on-chip and off-chip memory.
\item Similar to DASP, hardware design is synthesized using Vivado before being deployed to the Xilinx FPGA. The PS of the Zynq FPGA is used to control the dynamic streaming of model parameters FPGA (Xilinx- PYNQ Z2 in our case) into the select CNN layer.
\end{itemize}

 \begin{figure}[]
\centerline{\includegraphics[width=.5\textwidth]{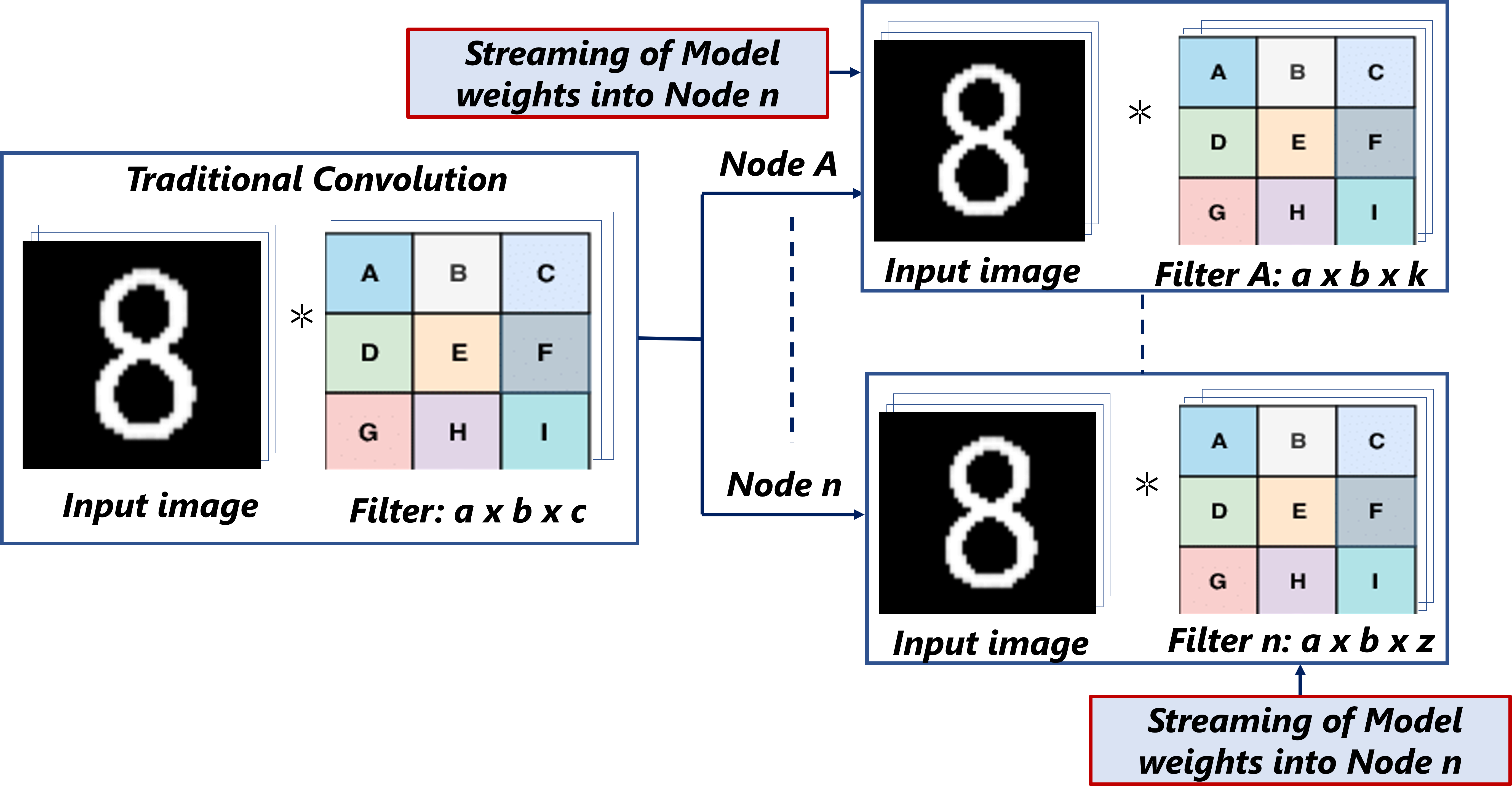}}
\caption{The traditional convolution operation is divided over several nodes in a resource constrained environment to achieve scalable dynamic CNN inference}
\label{image13}
\end{figure}

 \begin{figure}[ht]
\centerline{\includegraphics[width=.5\textwidth]{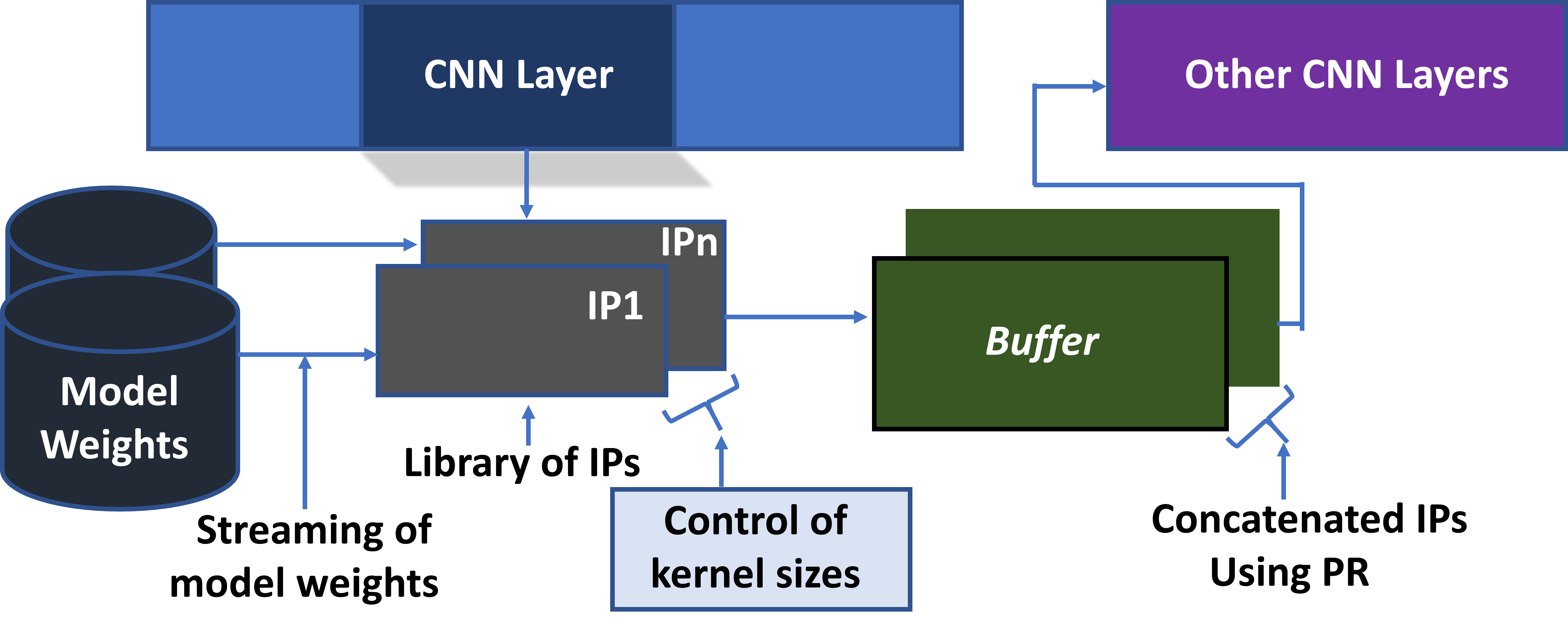}}
\caption{SDCI Operation in a CNN layer with large parameters: Following HLS design, multiple IPs with different kernel sizes could be generated, and their output could be cascaded to perform image classification using PR.}
\label{image8}
\end{figure}

\subsection{Scalable Dynamic CNN Inference (SDCI) using Library-Based PR} \label{SectionC}
\begin{figure}[]
\centerline{\includegraphics[width=.45\textwidth]{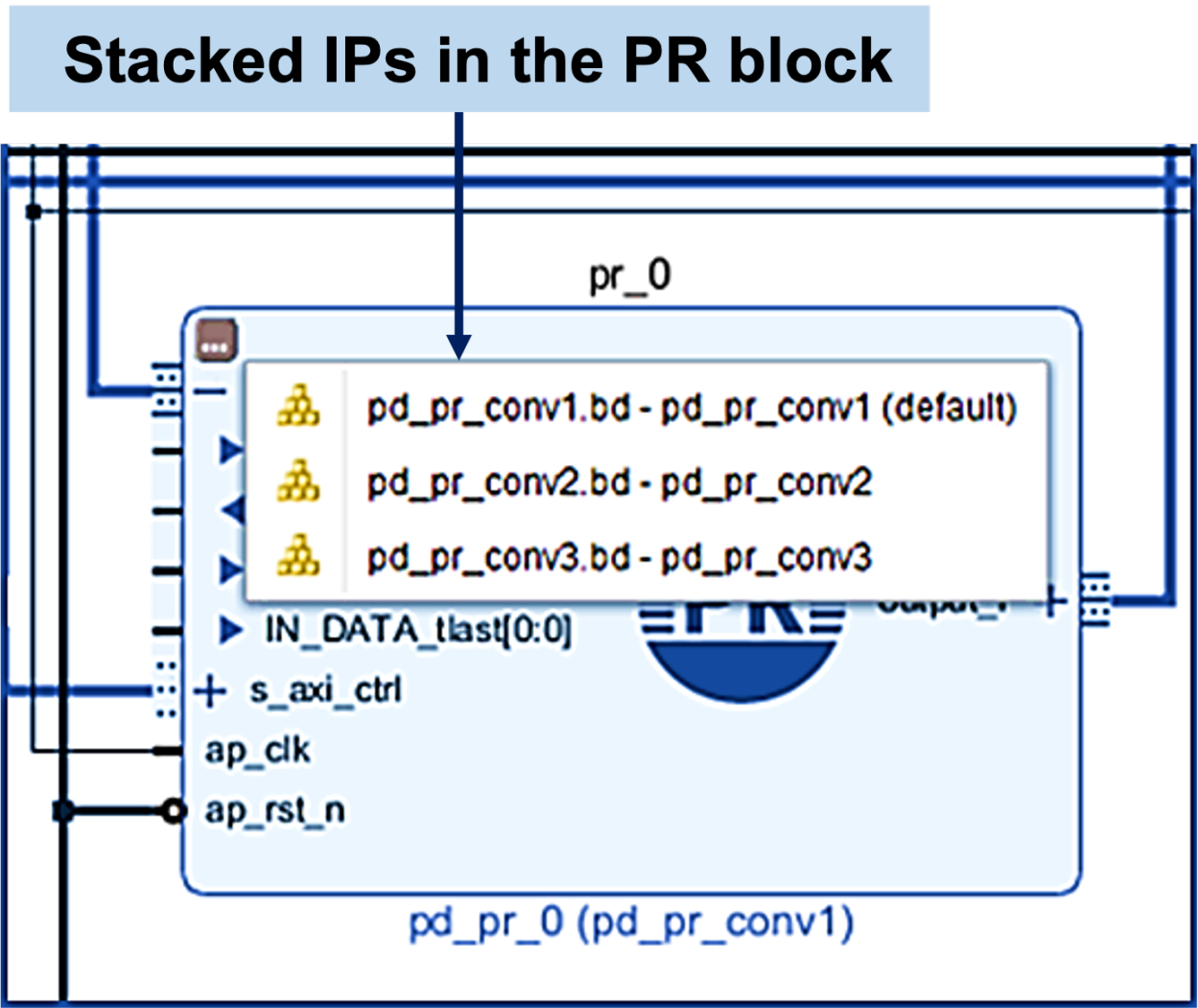}}
\caption{Leveraging partial reconfiguration to concatenate IPs generated by HLS during block design: The reconfigurable block allows for distributed convolution and the combining of fragmented IPs generated following HLS.}
\label{image10}
\end{figure}

The SDCI methodology is described using the below as described in Fig. \ref{image20}
\begin{itemize}
\item As is the case with other designs model parameters are obtained and saved in the PS after training the CNN model. 
\item To achieve dynamic resource allocation, one or more CNN layers can be designed dynamically using partial reconfiguration (PR) techniques. The rationale behind this is as follows. If the FPGA's resources are limited and it cannot accommodate the computation or does not have enough memory to perform the operation, it can be split into two or more operations. Having a library of IPs pre-synthesized and dynamically allocated on a FPGA can allow the FPGA designer to design any dimension of operation divided into multiple chunks. Fig. \ref{image13} shows one such example where a convolution operation is split into multiple small chunk of operation and eventually cascading of all these small operations can lead to complete result as depicted in Fig. \ref{image8}.  
\item In the hardware design, we prove the concept for implementing and adjusting model parameters on-the-fly by leveraging a library of pre-synthesized IPs, which are utilized dynamically via PR capability of FPGA design. Model parameters are adjusted on-the-fly to accommodate resource constrained nodes on an edge intelligent device (step 6- step 7). The IPs obtained for each of the node operation is combined using the Vivado's partial reconfiguration suite as shown in Fig. \ref{image10}. 
\end{itemize}

\section{Experimental Setup, Results and Discussion}

\begin{figure}[]
\centerline{\includegraphics[width=.45\textwidth]{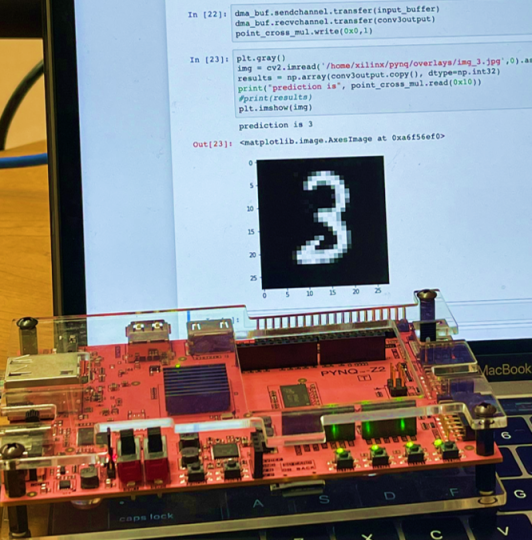}}
\caption{Validation of the proposed SDIC to demonstrate that classification is still correct after dynamic resource allocation}
\label{image9}
\end{figure}

\begin{table}[]
\centering
\caption{Resources, accuracy, and inference time for each methodology: SDCI utilises a static IP (IP3), while IP1 and IP2 are dynamically supplied. Since only one of the IPs (IP1 or IP2) is present on the FPGA at any given time, the resources are provided as such.}
\begin{tabular}{|c|c|c|cc|}
\hline
\textbf{Metrics} &
  \textbf{DASP} &
  \textbf{DAPP} &
  \multicolumn{1}{c|}{\textbf{\begin{tabular}[c]{@{}c@{}}SDCI\\ (IP1 + IP3)\end{tabular}}} &
  \textbf{\begin{tabular}[c]{@{}c@{}}SDCI\\ (IP2 + IP3)\end{tabular}} \\ \hline
FFs                                                                    & 7925  & 5832 & \multicolumn{1}{c|}{7883}  & 7885  \\ \hline
LUTs                                                                   & 7902  & 6229 & \multicolumn{1}{c|}{10155} & 10034 \\ \hline
DSPs                                                                   & 69    & 59   & \multicolumn{1}{c|}{29}    & 29    \\ \hline
BRAM                                                                   & 28.50 & 27   & \multicolumn{1}{c|}{35}    & 31    \\ \hline
\begin{tabular}[c]{@{}c@{}}Validation\\  Accuracy (\%)\end{tabular}    & 92    & 86   & \multicolumn{2}{c|}{94}            \\ \hline
\begin{tabular}[c]{@{}c@{}}Inference time \\ on PL (secs)\end{tabular} & 9.17     & 6.78 & \multicolumn{1}{c|}{26.91} & 21.09 \\ \hline
\end{tabular}
\label{table1}
\end{table}

The Xilinx's Vivado HLS 2018.3 is used to design the mapped CNN IP during HLS and an IP is obtained after the synthesis. Xilinx's Vivado 2018.3 is used to make the block design and integrate the created IP with AXI-interconnects and a ZYNQ processor (FPGA PYNQ-Z2 with a clock frequency of 100MHz). In this work, we propose three methodologies for dynamically providing model parameters to CNN at runtime during image classification. The methodologies are implemented on LeNet-5 trained on MNIST dataset. Resource utilization (BRAM, DSP, flip-flops (FFs), look-up tables (LUTs), Latency) and classification accuracy are evaluated for each methodology.


Table \ref{table1} shows that resource utilization (BRAMs, FFs, LUTs and DSPs) in the first methodology (DASP), where model all parameters are provided at run-time for all the CNN layers such that the CNN can be utilized dynamically. DASP offers a unique technique for a trained model to classify dataset that it has not previously been trained on, as weights could be dynamically provided to the model at run time. We performed image classification and achieved a classification accuracy of 92\%. For the DAPP technique, we streamed in model parameters for Conv1 and Conv2 layer of the LeNet-5 architecture while keeping the parameters of other layers in the on-chip memory. DAPP is a memory-constrained alternative to DASP. Model layers with large parameters may not fit on smaller FPGAs, so DAPP is used by keeping some of the network parameters static (on-chip), while the larger parameters are provided dynamically. Table \ref{table1} showed that BRAMs, FFs, LUTs and DSPs utilization reduced, at the expense of reduced accuracy of 86\%, which can be attributed to the quantization loss. 

Lastly, we leveraged on the PR suite of Vivado to perform scalable dynamic CNN inference by dynamically allocating resources to CNN layers. In this methodology (SDCI), we divided the convolution operation in the Conv1 layer of the network into two dynamic IPs. Using the LeNet-5 model and MNIST dataset, the first convolution operation necessitates the use of 6x5x5 filters and an image size of 28x28. During HLS, a first convolution with four filters (4x5x5) was performed with model parameters streamed in at run-time, yielding an IP1, followed by a second convolution with two filters and model parameters streamed in at run-time, yielding another IP2. During the block design, partial reconfiguration is used to cascade the results two IPs generated from convolution operations of the two dynamic IPs (IP1, and IP2). Overall setup is shown in Fig. \ref{image9}. Classification accuracy using SDCI is up to 94\%. SDCI takes the maximum time since dynamic allocation of resources require replacement of IPs. However, proof of concept implementation for dynamic implementation of CNN is shown. Further optimization will be reported elsewhere.

\section{CONCLUSION}
In this paper, we propose methodologies to perform dynamic allocation of parameters to CNN in hardware. Model parameters are allocated to  CNN layers for image classification. This proof of concept shows that model parameters could be stored in off-chip without compromising model accuracy. Consequently, to reduce bandwidth required to stream in model parameters, we prove that model layers with enormous parameter requirement could be streamed in during run time while model parameters of other smaller CNN layers could be stored on-chip without accuracy compromise. Lastly, we provide a proof of concept design technique and implementation for  performing dynamic allocation of resources in resource constrained environments by leveraging PR with library of IPs. Our experimental results show complete methodology with almost no accuracy loss and reasonable inference time on programming logic (PL).

\bibliographystyle{IEEEtran}
\bibliography{reference.bib}
\end{document}